\pdfoutput=1

\documentclass[11pt]{article}

\usepackage{EMNLP2023}

\usepackage{times}
\usepackage{latexsym}

\usepackage[T1]{fontenc}

\usepackage[utf8]{inputenc}

\usepackage{microtype}

\usepackage{inconsolata}

\usepackage{amsmath}
\usepackage{inconsolata}
\usepackage{hyperref}
\usepackage{paralist}
\usepackage{booktabs}
\usepackage{tikz} 
\usepackage{amssymb}
\usepackage{graphicx}
\usepackage{multirow}
\usepackage{multicol}
\usepackage{tabularx}
\usepackage{tikz}
\usepackage{array} 
\usepackage{subcaption}
\usepackage{float}
\usepackage{bm}
\usepackage{ulem}

\DeclareMathOperator{\newconsistency}{RC}
\DeclareMathOperator{\oldconsistency}{consistency}
\newcommand\sect[1]{\S\ref{#1}}

\newcommand{\orangecircle}{%
    \begin{tikzpicture}[baseline=-0.5ex] 
        \fill[orange] (0,0) circle (0.3cm);
    \end{tikzpicture}%
} 
\newcommand{\whitecircle}{%
    \begin{tikzpicture}[baseline=-0.5ex] 
        \fill[white] (0,0) circle (0.3cm);
    \end{tikzpicture}%
}

\title{How Much Consistency Is Your Accuracy Worth?}

\author{Jacob K.\ Johnson \and Ana Marasovi\'c\\
Kahlert School of Computing \\
University of Utah \\
\texttt{\{jacob.k.johnson,ana.marasovic\}@utah.edu}}

\begin{document}
\maketitle
\normalem
\begin{abstract}
Contrast set consistency is a robustness measurement that evaluates the rate at which a model correctly responds to all instances in a bundle of minimally different examples relying on the same knowledge. %
To draw additional insights, we propose to complement consistency with \emph{relative consistency}\,---\,the probability that an equally accurate model would surpass the consistency of the proposed model, given a distribution over possible consistencies. %
Models with 100\% relative consistency have reached a consistency peak for their accuracy. %
We reflect on prior work that reports consistency in contrast sets and observe that relative consistency can alter the assessment of a model's consistency compared to another. %
We anticipate that our proposed measurement and insights will influence future studies aiming to promote consistent behavior in models.
\end{abstract}

\section{Introduction}

Annotators introduce data shortcuts that allow models to solve tasks in unintended ways \cite{gururangan-etal-2018-annotation}. %
In response, it has been proposed to measure whether a model correctly responds to a bundle (or a \emph{contrast set}) of slightly modified instances that rely on the same knowledge \cite{gardner-etal-2020-evaluating, DBLP:conf/iclr/KaushikHL20}. %
The rate at which a model accomplishes this is termed \emph{consistency}. %
We propose an additional measurement\,---\,\emph{relative consistency}\,---\,that facilitates discussion about achievable consistency scores, enabling a more nuanced comparison.

\renewcommand{\arrayrulewidth}{1.5pt}

\newcolumntype{+}{!{\vrule width 1.5pt}}

\begin{table}[!ht]
    \centering
    \begin{subtable}{\linewidth} 
    \centering
    \begin{tabular}{+c+c+c+c+c+}
    \hline
    \orangecircle & \orangecircle & \whitecircle & \whitecircle & \whitecircle \\
    \hline
    \orangecircle & \orangecircle & \whitecircle & \whitecircle & \whitecircle \\
    \hline
    \end{tabular}%
    \caption{Accuracy=4/10, Consistency=2/5, RelConsistency=$100\%$}
    \label{tab:motivation_1}
    \end{subtable}
    \vspace{2mm}
    \begin{subtable}{\linewidth} 
    \centering
    \begin{tabular}{|c|c|c|c|c|}
    \hline
    \orangecircle & \orangecircle & \whitecircle & \whitecircle & \whitecircle \\
    \hline
    \orangecircle & \orangecircle & \orangecircle & \orangecircle & \orangecircle \\
    \hline
    \end{tabular}%
    \caption{Accuracy=7/10, Consistency=2/5, RelConsistency=$66.7\%$}
        \label{tab:motivation_2}
    \end{subtable}
    \vspace{2mm}
    \begin{subtable}{\linewidth} 
    \centering
    \begin{tabular}{|c|c|c|c|c|}
    \hline
    \orangecircle & \orangecircle & \orangecircle & \whitecircle & \whitecircle \\
    \hline
    \orangecircle & \orangecircle & \orangecircle & \orangecircle & \whitecircle \\
    \hline
    \end{tabular}
    \caption{Accuracy=7/10, Consistency=3/5, RelConsistency=$100\%$}
        \label{tab:motivation_3}
    \end{subtable}
    \vspace{2mm}
    \begin{subtable}{\linewidth} 
    \centering
    \begin{tabular}{|c|c|c|c|c|}
    \hline
    \orangecircle & \orangecircle & \orangecircle & \whitecircle & \whitecircle \\
    \hline
    \orangecircle & \orangecircle & \orangecircle & \orangecircle & \orangecircle \\
    \hline
    \end{tabular}%
    \caption{Accuracy=8/10, Consistency=3/5, RelConsistency=$88.9\%$}
        \label{tab:motivation_4}
    \end{subtable}
    \vspace{2mm}
    \begin{subtable}{\linewidth} 
    \centering
    \begin{tabular}{|c|c|c|c|c|}
    \hline
    \orangecircle & \orangecircle & \orangecircle & \orangecircle & \whitecircle \\
    \hline
    \orangecircle & \orangecircle & \orangecircle & \orangecircle & \whitecircle \\
    \hline
    \end{tabular}
    \caption{Accuracy=8/10, Consistency=4/5, RelConsistency=$100\%$}
        \label{tab:motivation_5}
    \end{subtable}
\caption{Tables depict a dataset of 10 examples, where each column showcases a bundle of an original instance paired with its perturbed version.  \begin{tikzpicture}[baseline=-0.5ex] \fill[orange] (0,0) circle (0.2cm); \end{tikzpicture} denotes that the instance is correctly predicted by a model. The relative consistency is the measurement we propose to complement the standard consistency.}
\label{tab:motivation}
\end{table}

To demonstrate why this is desired, consider situations that are illustrated in Table \ref{tab:motivation}. 
Both \ref{tab:motivation_1}--\ref{tab:motivation_2} correctly solve two bundles, i.e.,  have the same consistency. %
\ref{tab:motivation_2} solves three additional instances but in a way that does not promote consistency; \ref{tab:motivation_3} shows that a higher consistency can be gained with the same accuracy. %
In contrast, although \ref{tab:motivation_1} is less accurate, everything it handled was done consistently, and higher consistency cannot be achieved with the same accuracy. %
This analysis sheds light on an upside of \ref{tab:motivation_1} and a limitation of \ref{tab:motivation_2} that might go unnoticed if we solely compare accuracy/consistency. %
Let us turn to examples \ref{tab:motivation_4}. %
Although it represents a model with an improved consistency relative to \ref{tab:motivation_1}, we could have achieved better consistency for the same accuracy (see \ref{tab:motivation_5}).\footnote{Because this is a toy example, relative consistency is high, though not perfect, even in less-than-ideal  cases \ref{tab:motivation_2} and \ref{tab:motivation_4}.} %

Relative consistency (\sect{sec:relative_cons}) measures whether the consistency of our model would likely be outperformed by an equally accurate model, relative to the distribution of possible consistencies; see Eq.\ (\ref{eq:consistency_distr}). %
Specifically, it is the probability that our model's consistency is (in most cases) higher or equal to the consistency scores that are achievable with the same accuracy. %
If relative consistency is 100\% then our model is the most consistent it can be given its accuracy, as a more consistent, equally accurate model exists only with near-zero probability. %
In practice, the goal should be to increase the ``standard consistency'' while also achieving 100\% relative consistency. %

In light of this additional consistency metric, in \sect{sec:meta_analysis} we revisit the findings of three publications that report consistency as a metric for their evaluations and point out some additional conclusions we might draw from these reported consistencies. %
Our code is available at \url{https://github.com/jacobkj314/relative-consistency}.

\section{Relative Consistency}
\label{sec:relative_cons}

We first introduce background terminology (\sect{sec:background}), then derive elements we need for defining relative consistency: (i) achievable consistency scores for a given accuracy (\sect{sec:achievable_scores}) and (ii) a distribution over achievable consistency scores (\ref{sec:distribution}).

\subsection{Background}
\label{sec:background} 

A \emph{contrast set} or \emph{bundle} is a set of minimally different instances that might admit different answers, thus testing a model across/near its decision boundary.\footnote{Sometimes ``contrast set'' is used to refer to contrastive instances only (without the original ones).} %
For example, these two HotpotQA instances \cite{yang-etal-2018-hotpotqa} represent a contrast set: 
\begin{compactitem}
    \item Q: Is the Marsilea or the Brabejum the genus of \textbf{more} individual species of plants? A: Marsilea  
    \item Q: Is the Marsilea or the Brabejum the genus of \textbf{less} individual species of plants? A: Brabejum
\end{compactitem}
The model is required to answer both of them correctly to be considered consistent in that bundle. %
Evaluation with contrast sets makes it harder for simple and inadequate models to perform highly (e.g, a model that has just learned a spurious correlation between the word ``Marsilea'' and ``more''). %
Related studies construct bundles of paraphrases that have the same, not contrastive, labels \cite{elazar-etal-2021-measuring}. %

The term \emph{consistency} is overloaded in NLP and refers to different concepts \cite{li-etal-2019-logic,jang-etal-2022-becel,DBLP:conf/iclr/0002WSLCNCZ23}. %
In this work, we study \emph{contrast set consistency} defined as the proportion of bundles where a model accurately labels every instance in a bundle: 
\begin{equation}
    \oldconsistency=\frac{|B \in \mathcal{B}: \forall x \in B, y_p(x)=y(x)|}{|\mathcal{B}|},
\end{equation}
where $\mathcal{B}$ is a set of all bundles of related instances in a given dataset, $x$ is an example, $y_p(x)$ is the predicted label for $x$, and $y(x)$ is its gold label. %

\begin{figure*}[t]
    \centering
    \begin{subfigure}{0.48\textwidth}
        \includegraphics[width=\textwidth]{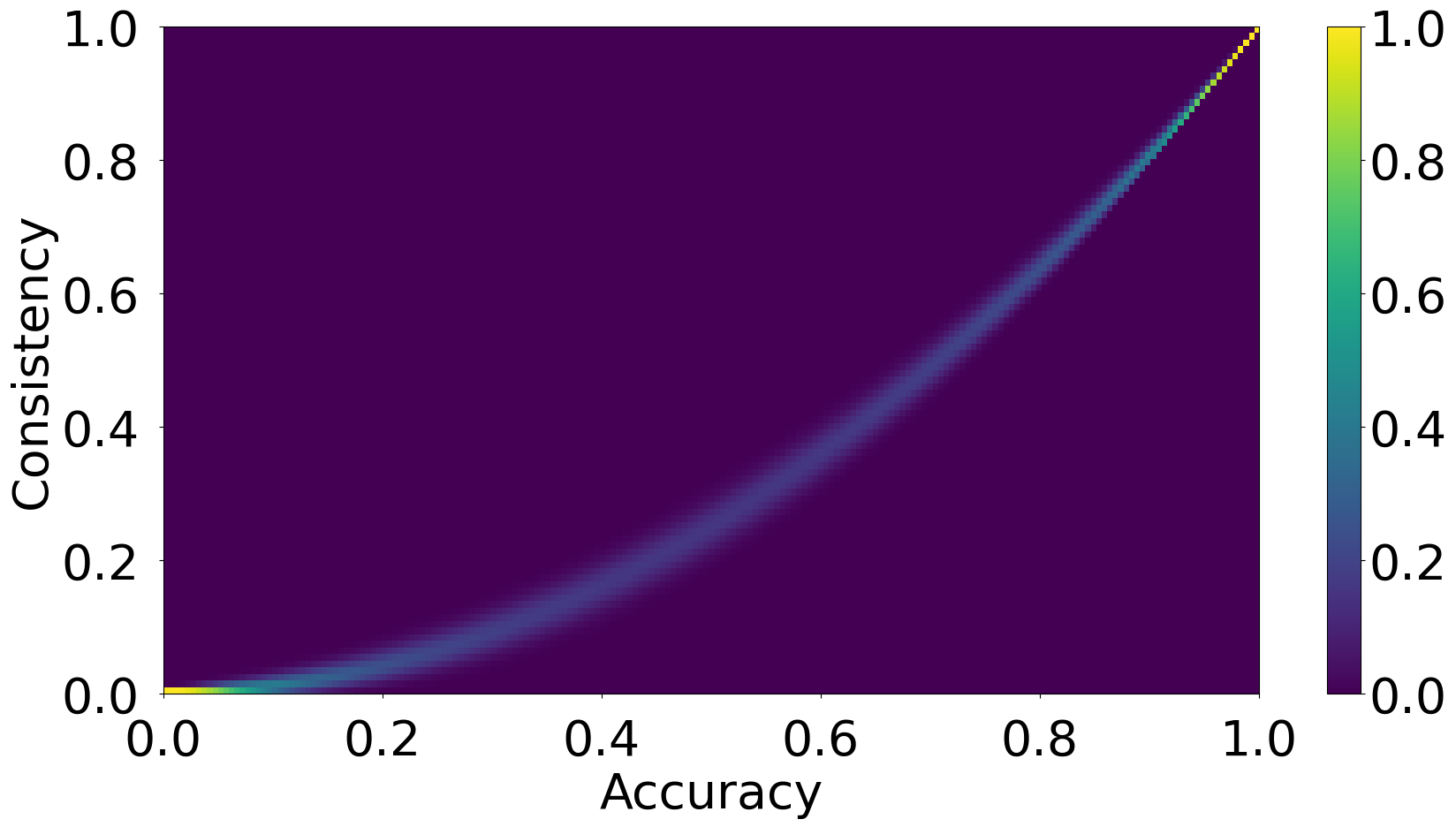}
        \caption{Distributions of consistency scores.}
        \label{fig:c_distribution_a}
    \end{subfigure}
    \hfill
    \begin{subfigure}{0.48\textwidth} 
        \includegraphics[width=\textwidth]{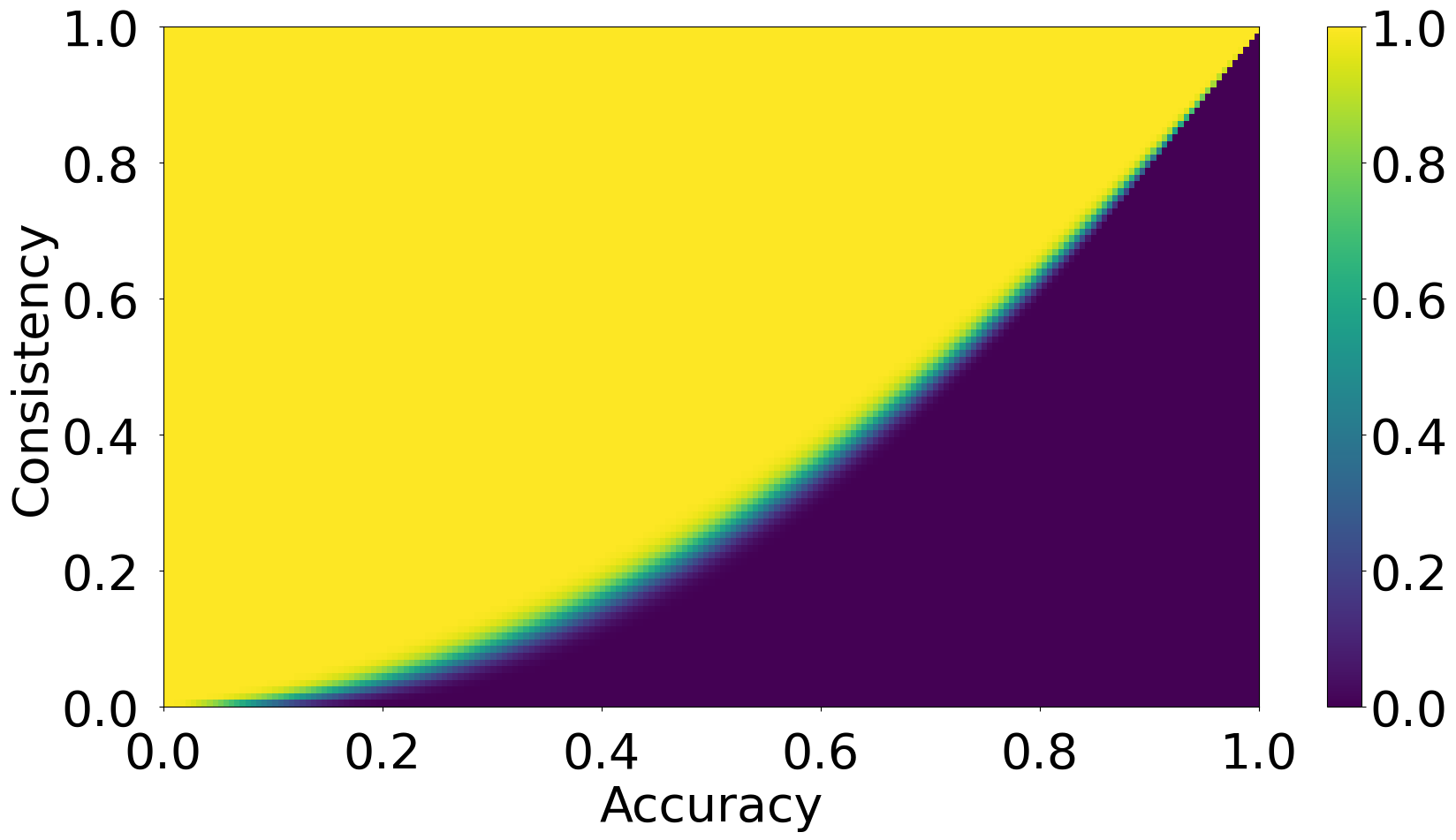}
        \caption{Relative consistency scores.} 
        \label{fig:rc}
    \end{subfigure}
    \caption{On the left is a heatmap of distributions of consistency at each accuracy for 100 bundles of 2 instances: each vertical slice corresponds to a separate distribution of different consistencies. 
    Fig.\ \ref{fig:distribution_log10} (Appendix) shows the $\log_{10}$ of this plot that better highlights the long tails of these distributions. On the right are relative consistency scores given a model's accuracy and consistency, i.e., the CDF of the figure on the left. Note that for a different number of bundles, these plots would look slightly different.}
    \label{fig:c_distribution}
\end{figure*}

\subsection{Achievable Consistency Scores}
\label{sec:achievable_scores}
        Consider a contrastive test set formed from $n$ original instances, plus a contrastive instance derived from each original instance by varying along some pertinent dimension. There are $2n+1$ possible accuracies $a$ that a model could achieve on this test set, namely $A = \{0 , 1 , \hdots , 2n-1 , 2n\}$.\footnote{While accuracy is typically denoted as a proportion of correct instances, reporting absolute numbers simplifies our notation. It is easy to translate a quantity $a$ to a corresponding proportion $\alpha$ via the identity $a = 2n\alpha$, while a consistency quantity $c$ relates to the consistency proportion $\gamma$ via $c = n\gamma$.} 
        Similarly, there are $n+1$ possible consistencies $c$ that a model could achieve, namely $C = \{0 , 1 , \hdots , n-1 , n\}$. 
        
        Furthermore, for a given accuracy $a \in A$, only a subset $C_a \subseteq C$ of consistencies is achievable. Trivially, for $a = 0$, $C_a = \{0\}$ (because a model cannot consistently respond to a bundle without correctly responding to at least the instances within that bundle)
        and for $a = 2n$, $C_a = \{n\}$ (because a model that correctly responds to all instances has also consistently responded to all the bundles those instances comprise). $C_a$  can then be defined in terms of $n$ and $a$:
            \begin{equation}
                C_a = \{c \in C : c_{min}^{(a)} \leq c \leq c_{max}^{(a)} \}
            \end{equation}
            where $c_{min}^{(a)}$ and $c_{max}^{(a)}$ are defined as:
            \begin{align}
                c_{min}^{(a)} &= 
                    \begin{cases}
                        0 & \text{if } a \leq n \\
                        a-n & \text{if } a > n
                    \end{cases}\\
               c_{max}^{(a)} &= \left\lfloor \frac{a}{2} \right\rfloor
            \end{align}
        Intuitively, if  $a \leq n$ then it is possible that all bundles have one of their constituent instances incorrectly answered, in which case, $c_{min}^{(a)} = 0$. However, if $a > n$, then at least $a - n > 0$ of bundles must be fully correctly answered. Indeed, for a bundle to be inconsistent at least one item must be incorrectly answered, so for a given $a$, the number of incorrect items is $2n-a$. Thus, at most $2n-a$ bundles can be inconsistent, and $c_{min}^{(a)} = n-(2n-a)=n-2n+a=a-n$. 

        The definition of $c_{max}^{(a)}$ follows from the observation that a maximally consistent model will consistently respond to the maximum number of bundles for which it is possible that both instances are correctly answered, and that equals $\left\lfloor \frac{a}{2} \right\rfloor$.

    \subsection{Distribution of Achievable Consistencies}
    \label{sec:distribution}
        Given an accuracy $a$, we construct a distribution of achievable consistencies $c \in C_a$ with:
        \begin{equation}
            \mathbb{P}(c | a) = \frac{m(c, a)}{M(a)}
            \label{eq:consistency_distr}
        \end{equation}
        where $M(a)$ is the number of ways a model can achieve accuracy $a$ and is given by:
        \begin{equation}
            M(a) = {2n \choose a}
        \end{equation}
        because there are $2n$ total instances, of which any $a$ might be the ones to which a model correctly responds.\footnote{It is possible to consider consistency to be the more underlying property of a model's behavior and compute a distribution over possible accuracies in the range $\left[2c,2n-n+c\right]$. The corresponding accuracy by consistency distributions could then be computed given the above-defined consistency by accuracy distributions.} 
        $m(c, a)$ represents the number of ways a model can achieve accuracy $a$ and consistency $c$, and is given by:
        \begin{equation}
            m(c, a) = {n \choose c}{{n-c} \choose {a-2c}} {2^{a-2c}} 
            \label{eq:m}
        \end{equation}
        where: 
        \begin{compactitem}
            \item ${n \choose c}$ corresponds to the number of ways that $c$ consistent bundles can be selected from $n$,
            \item ${{n-c} \choose {a-2c}}$ corresponds to the number of ways the remaining $a-2c$ accurate instances can be distributed across the remaining $n-c$ bundles, giving each selected bundle only one correct instance (to avoid creating an additional consistent bundle),
            \item $2^{a-2c}$ represents the number of ways that these partially correct bundles could have either instance correct.
        \end{compactitem}        
        
        Using this, we can calculate $m(c,a)$ and $M(a)$ across all values of $c$ and $a$ for reasonable sizes of $n$. These distributions can be extended for bundle sizes above 2; see formulas in Appendix \ref{appendix:b}. Figure \ref{fig:c_distribution_a} shows the distributions of consistency scores for a dataset with 100 bundles of 2 instances.

        Note that this distribution is not uniform for different consistencies at a given accuracy. %
        There will be some consistencies that have more ways to be achieved for a given accuracy. %
        This is why the formula $m(c,a)$ is crucial to the computation of relative consistency that comes next.

        This formulation assumes that all instances are equally difficult which 
        is known to not be the case in practice \cite{swayamdipta-etal-2020-dataset}. %
        It also disregards any inductive biases of models/datasets that could skew the distribution.

\paragraph{Relative Consistency}

    We measure the tendency to be consistent exhibited by a model that achieved accuracy $a$ and consistency $c$ on a contrastive set by computing the cumulative probability distribution over achievable consistencies in $C_a$ up to $c$: %

        \begin{equation}
            \newconsistency(c,a) = \sum_{\substack{c_i \in C_a \\ c_i \leq c}}{\mathbb{P}(c_i|a)}
        \end{equation}

    Intuitively, $\newconsistency(c, a)$ indicates how likely the model's consistency is to outperform an equally accurate model relative to the distribution of achievable consistencies defined in (\ref{eq:consistency_distr}). %
    This allows us to quantify whether model consistency is below, at, or above chance, given its accuracy. 
    In a good case, $\newconsistency$ is high, meaning that it is unlikely that an equally accurate model will have higher consistency. %
    Alternatively, if $\newconsistency$ is low, then it is likely that an equally accurate model will have higher consistency (which is unwanted).

Although other measurements which contextualize consistency scores within a particular accuracy can be constructed\,---\,such as simply scaling the consistency between $c_{min}^{(a)}$ and $c_{max}^{(a)}$, or reporting the fraction of fully consistent of those that are at least partly correct\,---\,these approaches lack the probabilistic interpretation underlying $\newconsistency$. %
\sect{sec:example}--\ref{sec:meta_analysis} highlight circumstances in which this probabilistic interpretation is useful, and Appendix \ref{appendix:other_distributions} compares the score distributions obtained via these measurements to the score distributions obtained via $\newconsistency$.

    \section{Analysis with Simulated Contrastive Set}
        \label{sec:example}
    
    Suppose you evaluate a model on a contrastive test set containing 100 bundles of 2 instances. %
    The distribution of consistencies 
    for this dataset is shown in Figure \ref{fig:c_distribution_a}, with the CDF of that distribution (corresponding to the $\newconsistency$ score) in Figure \ref{fig:rc}.
    
    Note that the highest-density region of the distribution moves upward as accuracy increases, and takes up only a very thin band. %
    This means that, for a given accuracy, there is generally little room for improvement in consistency. This can be useful when discussing results: if a particular training approach yields a 5\% improvement in consistency for an equally accurate model, that represents a substantial change in how the model tends to respond to inputs.

    It can still happen that improving accuracy and consistency decreases relative consistency. %
    As an example, consider comparing a model $M_1$, which achieves $a=130, c=45$ ($65\%$ accuracy, $45\%$ consistency) against a model $M_2$ with $a=150, c=55$ ($75\%$ accuracy, $55\%$ consistency). %
    Clearly, model $M_2$ is more desirable for practical uses, if we are just comparing one model to another, but if we are comparing two different training approaches, and want to know which induces a stronger tendency for consistent responses, then we would be interested to know that $M_1$ has $\newconsistency=93.0\%$, while $M_2$ has $\newconsistency=37.1\%$. %
    This insight, that one model is below chance consistency, while another is well above, is made possible by the probabilistic interpretation of $\newconsistency$.

\section{Meta-Analysis of Prior Work}
\label{sec:meta_analysis}
    In this section, we discuss results reported by prior works that conduct evaluation with contrast sets under the light of relative consistency. %

\begin{table}[t]
\centering
\resizebox{\columnwidth}{!}{
\begin{tabular}{lcccr}
\toprule
\textbf{Dataset} & \textbf{\#Bundles} & \textbf{Acc} & \textbf{Cons} & \bm{$\newconsistency$} \\ 
\midrule
UD Parsing       & 150        & 55.3        & 17.3          & $\sim$0.0 \\ 
PERSPECTRUM      & 217        & 88.0         & 78.8          & 97.6             \\
ROPES            & 974        & 40.1         & 17.6          & 97.8             \\
MC-TACO          & 646        & 26.0         & 8.0           & 95.2             \\ 
\bottomrule
\end{tabular}
}
\caption{
    Relative consistency scores computed for results reported in \citet{gardner-etal-2020-evaluating}. In the 3rd column, we report the average of ``Original Test'' (original only) and ``Contrast'' (contrastive only) columns in their Table 2. That is the accuracy, $a$, we use in calculations in \sect{sec:relative_cons}. Models with similar consistency (UD Parsing and ROPES) have different tendencies to respond consistently as revealed by their $\newconsistency$ scores. 
}
\label{tab:gardner_et_al}
\end{table}
\subsection{\citet{gardner-etal-2020-evaluating}}

They construct contrast sets for several common test sets by modifying a sample of the test set instances. %
They train a biaffine parser \cite{dozat2017deep} with ELMo embeddings \cite{peters-etal-2018-deep} for UD parsing (\citealp{zeldes2017gum}, \citealp{silveira-etal-2014-gold}, \citealp{sanguinetti2015parli}, \citealp{ahrenberg-2007-lines}), and RoBERTa \cite{liu2019roberta} for reading comprehension tasks: ROPES \cite{lin-etal-2019-reasoning}, and MC-TACO \cite{zhou-etal-2019-going} and stance prediction: PERSPECTRUM \cite{chen-etal-2019-seeing}. %
Table \ref{tab:gardner_et_al} shows the accuracy and consistency of these models for four of their contrast sets.\footnote{We exclude contrast sets that do not have the bundle size of 2. They report the accuracy of the original instances and contrastive instances separately, so to obtain the accuracy in the contrast set (that we need to calculate $\newconsistency$) we average those. In doing so, we assume that the accuracy of the full original test set is similar to the accuracy of the sample of original test set instances.}  
In the rightmost column, we report the relative consistency scores that we introduce. %
    
\paragraph{Analysis} We observe that the UD parsing and ROPES models have a similar consistency score (17.3 and 17.6). %
However, the UD parsing model's consistency has a near-zero chance to outperform an equally accurate model. %
On the other hand, the ROPES model is quite likely to do so. 
            
Additionally, relative consistency shows that models with low consistency could nonetheless have a large tendency to respond to bundles consistently.\footnote{Note that high relative consistency does not guarantee that such a model will continue to respond to bundles consistently with improved accuracy.}
We see this with the results for MC-TACO, which, despite only achieving 8.0\% consistency, is more consistent than an equally accurate model in 95.2\% of cases. 
Intuitively, this means that the above chance model has at least generalized well within the few cases to which it correctly responds.

\subsection{\citet{dua-etal-2021-learning}}

They investigate whether training approaches that consider a full bundle of related instances together,  instead of their constituent instances separately, improve consistency. %
Table \ref{tab:dua_et_al} shows their report results obtained with T5 \cite{raffel2020exploring} and the relative consistency scores we compute from their results, on the contrastive version of ROPES\,---\,a reading comprehension dataset for evaluating a model's ability to reason about  ``effects of the relationships in the background passage in the context of the situation''.
       
\begin{table}[t]
\centering
\resizebox{\columnwidth}{!}{
\begin{tabular}{lccc}
\toprule
\textbf{Loss}  & \textbf{Accuracy} & \textbf{Consistency} & \bm{$\newconsistency$} \\
\midrule
MLE                     & 65.7         & 52.1          & 100.0         \\
\quad \rotatebox[origin=c]{180}{$\Lsh$} +UL                  & 68.3         & 55.6          & 100.0         \\
\quad \rotatebox[origin=c]{180}{$\Lsh$} +CE                  & 76.6         & 64.7          & 100.0         \\
\bottomrule
\end{tabular}
}
\caption{A comparison of relative consistency scores computed from results report in \citet{dua-etal-2021-learning} (in ``Dev EM'' and ``Dev C'' columns in their Table 3). The number of bundles is 844. The unlikelihood (UL) and contrastive estimation (CE) objectives improved the accuracy and consistency over MLE, \emph{without decreasing relative consistency}. This is how consistency should be improved in this case. 
}
\label{tab:dua_et_al}
\end{table}
\begin{table*}[t]
\resizebox{\textwidth}{!}{
\begin{tabular}{llrrrrrrrrrrrr}
\toprule
\textbf{Size} & \textbf{Loss} & \textbf{Scope-Acc} & \textbf{Aff-Acc} & \textbf{Scope-Cons} & \textbf{Aff-Cons} & \textbf{Scope-RC} & \textbf{Aff-RC} \\
\midrule

\multirow{5}{*}{\rotatebox{90}{Large}} 
    & MLE &
        66.84 & 67.09 & \uwave{42.86} &                       42.35  & \uwave{17.10} &                       10.06  \\
    & CE &
        64.80 & 66.84 &                        40.31 & \uuline{43.37} &                       20.10  & \uuline{26.64} \\
    & $\lambda_1$MLE + $\lambda_2$CE\\
    & \quad \rotatebox[origin=c]{180}{$\Lsh$} $\lambda_1, \lambda_2=1.0,1.0$ & 
        66.33 & 68.11 & \uwave{42.86} & \uuline{44.39} & \uwave{30.43} & \uuline{19.37} \\
    & \quad \rotatebox[origin=c]{180}{$\Lsh$} $\lambda_1, \lambda_2=0.33,1.0$ &
        66.58 & 68.37 &                       \underline{43.37}  &                       44.90  &                       \underline{42.44}  &                       16.01  \\

\midrule

\multirow{5}{*}{\rotatebox{90}{3B}} 
    & MLE &
        74.23 & 76.79 & \emph{56.12} &                       60.71  & \emph{80.76} &                       88.68  \\
    & CE &
        74.23 & 77.55 & 56.12 & 61.73 & 80.76 & 92.03 \\
    & $\lambda_1$MLE + $\lambda_2$CE\\
    & \quad \rotatebox[origin=c]{180}{$\Lsh$} $\lambda_1, \lambda_2=0.33,1.0$ &
        74.23   &   77.04   &   56.12   &   60.71   &   80.76  &   88.68  \\
    & \quad \rotatebox[origin=c]{180}{$\Lsh$} $\lambda_1, \lambda_2=0.33,1.0$ &
        76.02 & 78.57 & \emph{58.67} &                       63.78  & \emph{79.32} &                       98.60  \\ 
        \bottomrule

\end{tabular}
}
\caption{Results of UnifiedQA-v2 \cite{khashabi2022unifiedqav2} on the CondaQA contrastive dataset, with the expectation that including the Contrastive Estimation (CE) objective would improve consistency, as in \citet{dua-etal-2021-learning}. $\newconsistency$ scores are reported here only for some of the edit dimensions in CondaQA; see Table \ref{tab:ravichander_et_al-appendix} for the rest.  
}
\label{tab:ravichander_et_al}
\end{table*}

\paragraph{Analysis } We observe that the baseline model trained with the maximum likelihood estimation (MLE) is already at ceiling performance in terms of its tendency to produce consistent responses (i.e., its $\newconsistency$ scores). %
Combining contrastive estimation \cite[CE;][]{smith-eisner-2005-contrastive}, or unlikelihood training \cite[UL;][]{DBLP:conf/iclr/WelleckKRDCW20}, with MLE not only improves the accuracy and consistency but also does so in a way that does not lower the relative consistency, which is desired. %
This emphasizes the effectiveness of these objectives.

\subsection{\citet{ravichander-etal-2022-condaqa}}

They introduce CondaQA, a contrastive dataset for studying reading comprehension models' effectiveness in reasoning about the implications of negation expressed in a given text. %
Each CondaQA instance comes with three minimally varied versions: one paraphrases the negation, another modifies what is negated (scope), and the last removes the negation. %
\citet{ravichander-etal-2022-condaqa} use UnifiedQA-v2 \cite{khashabi2022unifiedqav2} as a backbone model. %
We explore the factors that might influence the consistency of the large and 3B versions of this model: 
\begin{compactitem}
    \item The training objective: MLE, CE, or combined $\lambda_1$MLE+$\lambda_2$CE. 
    \item The choice of hyperparemeters $\lambda_1$ and $\lambda_2$ (with UnifiedQA-large).
\end{compactitem}
Table \ref{tab:ravichander_et_al} shows accuracy, consistency, and relative consistency we obtain for bundles where the original instance is paired with its: (i) \emph{scope}-edited version, and (ii) \emph{affirmative} version (without negation). %
In Table \ref{tab:ravichander_et_al-appendix} (Appendix), we also include the results with paraphrase-edits.

\paragraph{Analysis}    
An increase in consistency does not necessarily indicate a heightened tendency to consistently respond to bundles (unless the accuracy stays the same). %
Compare CE with 1MLE+1CE (double underlined, in the upper part of the table). %
In this case, by training with MLE and CE, affirmative consistency has gone up slightly, however, the model's chance of outperforming an equally accurate model dropped down from 26\% to 19\%. %
This is an example of a suboptimal way of improving consistency, and MLE+CE is not necessarily superior to the standalone CE in this case. %
A similar, but less pronounced, situation occurs when comparing MLE against .33MLE+1CE for scope consistency in the bottom part of the table (italicized).
            
Conversely, even if standard consistency has not improved, a model's tendency to consistently respond to bundles may have. %
For example, compare MLE with 1MLE+1CE for scope consistency in the upper part of the table (wavy underlined). %
In this case, scope accuracy lowered slightly but absolute scope consistency remained the same, leading to a large improvement in Scope-$\newconsistency$. %
This may suggest that additional CE loss resulted in the model unlearning a few individual instances without unlearning any complete bundles it had learned. %
Similarly, 0.33MLE+1CE scope consistency in the upper part of the table (underlined once) increased slightly but the scope relative consistency has increased notably. %
If we compared only consistency we would conclude that the choice of hyperparameters $\lambda_1,\lambda_2$ is not vital, where actually they can affect model consistency behavior as shown by relative consistency.

\section{Conclusion}
We introduce relative consistency, which complements standard contrast consistency by allowing an accuracy and consistency score pair to be examined to determine whether a higher consistency was possible with that accuracy. %
This facilitates the comparison of consistencies achieved by models that achieved different levels of accuracy. %
We show that relative consistency enriches conclusions we make about whether a model is more consistent than another, and occasionally even leads us to different takeaways. 

\section{Limitations}
    This mathematical model is based on a simplified version of contrastive datasets. %
    Contrastive datasets may have more than two edits for each original instance, which will result in a different distribution. %
    Although we provide formulas for distributions of arbitrary bundle size in Appendix \ref{appendix:b}, these distributions are less intuitive, more expensive to compute, and additionally have the drawback that, if a model achieves high pairwise $\newconsistency$ on two of the elements of the bundle, it is likely to achieve high bundle $\newconsistency$, even if the other elements of the test set do not achieve high pairwise $\newconsistency$. %
    In general, we recommend formulating questions of consistency in terms of bundles with one instance exhibiting a feature and the other instance lacking that feature.
    Moreover, contrastive datasets may include extra data that is not contrastive; e.g., CondaQA has a small number of bundles with a single instance because other instances in the bundle were filtered because they did not pass quality checks. %

    In \sect{sec:distribution}, we state the drawbacks of the distribution (\ref{eq:consistency_distr}). %
    Namely, we do not consider that the distribution might be skewed due to the varying example difficulty and other inherent properties of datasets and models. 

\section{Acknowledgements}
We thank anonymous reviewers for their thoughtful and constructive comments,  members of the UtahNLP group for helpful feedback, and Petar Baki\'c for proofreading our formulas. 

\bibliography{anthology,custom}

\begin{thebibliography}{25}
\expandafter\ifx\csname natexlab\endcsname\relax\def\natexlab#1{#1}\fi

\bibitem[{Ahrenberg(2007)}]{ahrenberg-2007-lines}
Lars Ahrenberg. 2007.
\newblock \href {https://aclanthology.org/W07-2441} {{L}in{ES}: An
  {E}nglish-{S}wedish parallel treebank}.
\newblock In \emph{Proceedings of the 16th Nordic Conference of Computational
  Linguistics ({NODALIDA} 2007)}, pages 270--273, Tartu, Estonia. University of
  Tartu, Estonia.

\bibitem[{Basili et~al.(2015)Basili, Bosco, Delmonte, Moschitti, and
  Simi}]{sanguinetti2015parli}
Roberto Basili, Cristina Bosco, Rodolfo Delmonte, Alessandro Moschitti, and
  Maria Simi. 2015.
\newblock \href {https://doi.org/10.1007/978-3-319-14206-7}
  {\emph{Harmonization and Development of Resources and Tools for Italian
  Natural Language Processing within the PARLI Project}}, volume 589.

\bibitem[{Chen et~al.(2019)Chen, Khashabi, Yin, Callison-Burch, and
  Roth}]{chen-etal-2019-seeing}
Sihao Chen, Daniel Khashabi, Wenpeng Yin, Chris Callison-Burch, and Dan Roth.
  2019.
\newblock \href {https://doi.org/10.18653/v1/N19-1053} {Seeing things from a
  different angle:discovering diverse perspectives about claims}.
\newblock In \emph{Proceedings of the 2019 Conference of the North {A}merican
  Chapter of the Association for Computational Linguistics: Human Language
  Technologies, Volume 1 (Long and Short Papers)}, pages 542--557, Minneapolis,
  Minnesota. Association for Computational Linguistics.

\bibitem[{Dozat and Manning(2017)}]{dozat2017deep}
Timothy Dozat and Christopher~D. Manning. 2017.
\newblock \href {http://arxiv.org/abs/1611.01734} {Deep biaffine attention for
  neural dependency parsing}.

\bibitem[{Dua et~al.(2021)Dua, Dasigi, Singh, and
  Gardner}]{dua-etal-2021-learning}
Dheeru Dua, Pradeep Dasigi, Sameer Singh, and Matt Gardner. 2021.
\newblock \href {https://doi.org/10.18653/v1/2021.emnlp-main.584} {Learning
  with instance bundles for reading comprehension}.
\newblock In \emph{Proceedings of the 2021 Conference on Empirical Methods in
  Natural Language Processing}, pages 7347--7357, Online and Punta Cana,
  Dominican Republic. Association for Computational Linguistics.

\bibitem[{Elazar et~al.(2021)Elazar, Kassner, Ravfogel, Ravichander, Hovy,
  Sch{\"u}tze, and Goldberg}]{elazar-etal-2021-measuring}
Yanai Elazar, Nora Kassner, Shauli Ravfogel, Abhilasha Ravichander, Eduard
  Hovy, Hinrich Sch{\"u}tze, and Yoav Goldberg. 2021.
\newblock \href {https://doi.org/10.1162/tacl_a_00410} {Measuring and improving
  consistency in pretrained language models}.
\newblock \emph{Transactions of the Association for Computational Linguistics},
  9:1012--1031.

\bibitem[{Gardner et~al.(2020)Gardner, Artzi, Basmov, Berant, Bogin, Chen,
  Dasigi, Dua, Elazar, Gottumukkala, Gupta, Hajishirzi, Ilharco, Khashabi, Lin,
  Liu, Liu, Mulcaire, Ning, Singh, Smith, Subramanian, Tsarfaty, Wallace,
  Zhang, and Zhou}]{gardner-etal-2020-evaluating}
Matt Gardner, Yoav Artzi, Victoria Basmov, Jonathan Berant, Ben Bogin, Sihao
  Chen, Pradeep Dasigi, Dheeru Dua, Yanai Elazar, Ananth Gottumukkala, Nitish
  Gupta, Hannaneh Hajishirzi, Gabriel Ilharco, Daniel Khashabi, Kevin Lin,
  Jiangming Liu, Nelson~F. Liu, Phoebe Mulcaire, Qiang Ning, Sameer Singh,
  Noah~A. Smith, Sanjay Subramanian, Reut Tsarfaty, Eric Wallace, Ally Zhang,
  and Ben Zhou. 2020.
\newblock \href {https://doi.org/10.18653/v1/2020.findings-emnlp.117}
  {Evaluating models{'} local decision boundaries via contrast sets}.
\newblock In \emph{Findings of the Association for Computational Linguistics:
  EMNLP 2020}, pages 1307--1323, Online. Association for Computational
  Linguistics.

\bibitem[{Gururangan et~al.(2018)Gururangan, Swayamdipta, Levy, Schwartz,
  Bowman, and Smith}]{gururangan-etal-2018-annotation}
Suchin Gururangan, Swabha Swayamdipta, Omer Levy, Roy Schwartz, Samuel Bowman,
  and Noah~A. Smith. 2018.
\newblock \href {https://doi.org/10.18653/v1/N18-2017} {Annotation artifacts in
  natural language inference data}.
\newblock In \emph{Proceedings of the 2018 Conference of the North {A}merican
  Chapter of the Association for Computational Linguistics: Human Language
  Technologies, Volume 2 (Short Papers)}, pages 107--112, New Orleans,
  Louisiana. Association for Computational Linguistics.

\bibitem[{Jang et~al.(2022)Jang, Kwon, and Lukasiewicz}]{jang-etal-2022-becel}
Myeongjun Jang, Deuk~Sin Kwon, and Thomas Lukasiewicz. 2022.
\newblock \href {https://aclanthology.org/2022.coling-1.324} {{BECEL}:
  Benchmark for consistency evaluation of language models}.
\newblock In \emph{Proceedings of the 29th International Conference on
  Computational Linguistics}, pages 3680--3696, Gyeongju, Republic of Korea.
  International Committee on Computational Linguistics.

\bibitem[{Kaushik et~al.(2020)Kaushik, Hovy, and
  Lipton}]{DBLP:conf/iclr/KaushikHL20}
Divyansh Kaushik, Eduard~H. Hovy, and Zachary~Chase Lipton. 2020.
\newblock \href {https://openreview.net/forum?id=Sklgs0NFvr} {Learning the
  difference that makes {A} difference with counterfactually-augmented data}.
\newblock In \emph{8th International Conference on Learning Representations,
  {ICLR} 2020, Addis Ababa, Ethiopia, April 26-30, 2020}. OpenReview.net.

\bibitem[{Khashabi et~al.(2022)Khashabi, Kordi, and
  Hajishirzi}]{khashabi2022unifiedqav2}
Daniel Khashabi, Yeganeh Kordi, and Hannaneh Hajishirzi. 2022.
\newblock \href {http://arxiv.org/abs/2202.12359} {Unifiedqa-v2: Stronger
  generalization via broader cross-format training}.

\bibitem[{Li et~al.(2019)Li, Gupta, Mehta, and Srikumar}]{li-etal-2019-logic}
Tao Li, Vivek Gupta, Maitrey Mehta, and Vivek Srikumar. 2019.
\newblock \href {https://doi.org/10.18653/v1/D19-1405} {A logic-driven
  framework for consistency of neural models}.
\newblock In \emph{Proceedings of the 2019 Conference on Empirical Methods in
  Natural Language Processing and the 9th International Joint Conference on
  Natural Language Processing (EMNLP-IJCNLP)}, pages 3924--3935, Hong Kong,
  China. Association for Computational Linguistics.

\bibitem[{Lin et~al.(2019)Lin, Tafjord, Clark, and
  Gardner}]{lin-etal-2019-reasoning}
Kevin Lin, Oyvind Tafjord, Peter Clark, and Matt Gardner. 2019.
\newblock \href {https://doi.org/10.18653/v1/D19-5808} {Reasoning over
  paragraph effects in situations}.
\newblock In \emph{Proceedings of the 2nd Workshop on Machine Reading for
  Question Answering}, pages 58--62, Hong Kong, China. Association for
  Computational Linguistics.

\bibitem[{Liu et~al.(2019)Liu, Ott, Goyal, Du, Joshi, Chen, Levy, Lewis,
  Zettlemoyer, and Stoyanov}]{liu2019roberta}
Yinhan Liu, Myle Ott, Naman Goyal, Jingfei Du, Mandar Joshi, Danqi Chen, Omer
  Levy, Mike Lewis, Luke Zettlemoyer, and Veselin Stoyanov. 2019.
\newblock \href {http://arxiv.org/abs/1907.11692} {Roberta: A robustly
  optimized bert pretraining approach}.

\bibitem[{Peters et~al.(2018)Peters, Neumann, Iyyer, Gardner, Clark, Lee, and
  Zettlemoyer}]{peters-etal-2018-deep}
Matthew~E. Peters, Mark Neumann, Mohit Iyyer, Matt Gardner, Christopher Clark,
  Kenton Lee, and Luke Zettlemoyer. 2018.
\newblock \href {https://doi.org/10.18653/v1/N18-1202} {Deep contextualized
  word representations}.
\newblock In \emph{Proceedings of the 2018 Conference of the North {A}merican
  Chapter of the Association for Computational Linguistics: Human Language
  Technologies, Volume 1 (Long Papers)}, pages 2227--2237, New Orleans,
  Louisiana. Association for Computational Linguistics.

\bibitem[{Raffel et~al.(2020)Raffel, Shazeer, Roberts, Lee, Narang, Matena,
  Zhou, Li, and Liu}]{raffel2020exploring}
Colin Raffel, Noam Shazeer, Adam Roberts, Katherine Lee, Sharan Narang, Michael
  Matena, Yanqi Zhou, Wei Li, and Peter~J. Liu. 2020.
\newblock \href {http://arxiv.org/abs/1910.10683} {Exploring the limits of
  transfer learning with a unified text-to-text transformer}.

\bibitem[{Ravichander et~al.(2022)Ravichander, Gardner, and
  Marasovic}]{ravichander-etal-2022-condaqa}
Abhilasha Ravichander, Matt Gardner, and Ana Marasovic. 2022.
\newblock \href {https://aclanthology.org/2022.emnlp-main.598} {{CONDAQA}: A
  contrastive reading comprehension dataset for reasoning about negation}.
\newblock In \emph{Proceedings of the 2022 Conference on Empirical Methods in
  Natural Language Processing}, pages 8729--8755, Abu Dhabi, United Arab
  Emirates. Association for Computational Linguistics.

\bibitem[{Silveira et~al.(2014)Silveira, Dozat, de~Marneffe, Bowman, Connor,
  Bauer, and Manning}]{silveira-etal-2014-gold}
Natalia Silveira, Timothy Dozat, Marie-Catherine de~Marneffe, Samuel Bowman,
  Miriam Connor, John Bauer, and Chris Manning. 2014.
\newblock \href
  {http://www.lrec-conf.org/proceedings/lrec2014/pdf/1089_Paper.pdf} {A gold
  standard dependency corpus for {E}nglish}.
\newblock In \emph{Proceedings of the Ninth International Conference on
  Language Resources and Evaluation ({LREC}'14)}, pages 2897--2904, Reykjavik,
  Iceland. European Language Resources Association (ELRA).

\bibitem[{Smith and Eisner(2005)}]{smith-eisner-2005-contrastive}
Noah~A. Smith and Jason Eisner. 2005.
\newblock \href {https://doi.org/10.3115/1219840.1219884} {Contrastive
  estimation: Training log-linear models on unlabeled data}.
\newblock In \emph{Proceedings of the 43rd Annual Meeting of the Association
  for Computational Linguistics ({ACL}{'}05)}, pages 354--362, Ann Arbor,
  Michigan. Association for Computational Linguistics.

\bibitem[{Swayamdipta et~al.(2020)Swayamdipta, Schwartz, Lourie, Wang,
  Hajishirzi, Smith, and Choi}]{swayamdipta-etal-2020-dataset}
Swabha Swayamdipta, Roy Schwartz, Nicholas Lourie, Yizhong Wang, Hannaneh
  Hajishirzi, Noah~A. Smith, and Yejin Choi. 2020.
\newblock \href {https://doi.org/10.18653/v1/2020.emnlp-main.746} {Dataset
  cartography: Mapping and diagnosing datasets with training dynamics}.
\newblock In \emph{Proceedings of the 2020 Conference on Empirical Methods in
  Natural Language Processing (EMNLP)}, pages 9275--9293, Online. Association
  for Computational Linguistics.

\bibitem[{Wang et~al.(2023)Wang, Wei, Schuurmans, Le, Chi, Narang, Chowdhery,
  and Zhou}]{DBLP:conf/iclr/0002WSLCNCZ23}
Xuezhi Wang, Jason Wei, Dale Schuurmans, Quoc~V. Le, Ed~H. Chi, Sharan Narang,
  Aakanksha Chowdhery, and Denny Zhou. 2023.
\newblock \href {https://openreview.net/pdf?id=1PL1NIMMrw} {Self-consistency
  improves chain of thought reasoning in language models}.
\newblock In \emph{The Eleventh International Conference on Learning
  Representations, {ICLR} 2023, Kigali, Rwanda, May 1-5, 2023}. OpenReview.net.

\bibitem[{Welleck et~al.(2020)Welleck, Kulikov, Roller, Dinan, Cho, and
  Weston}]{DBLP:conf/iclr/WelleckKRDCW20}
Sean Welleck, Ilia Kulikov, Stephen Roller, Emily Dinan, Kyunghyun Cho, and
  Jason Weston. 2020.
\newblock \href {https://openreview.net/forum?id=SJeYe0NtvH} {Neural text
  generation with unlikelihood training}.
\newblock In \emph{8th International Conference on Learning Representations,
  {ICLR} 2020, Addis Ababa, Ethiopia, April 26-30, 2020}. OpenReview.net.

\bibitem[{Yang et~al.(2018)Yang, Qi, Zhang, Bengio, Cohen, Salakhutdinov, and
  Manning}]{yang-etal-2018-hotpotqa}
Zhilin Yang, Peng Qi, Saizheng Zhang, Yoshua Bengio, William Cohen, Ruslan
  Salakhutdinov, and Christopher~D. Manning. 2018.
\newblock \href {https://doi.org/10.18653/v1/D18-1259} {{H}otpot{QA}: A dataset
  for diverse, explainable multi-hop question answering}.
\newblock In \emph{Proceedings of the 2018 Conference on Empirical Methods in
  Natural Language Processing}, pages 2369--2380, Brussels, Belgium.
  Association for Computational Linguistics.

\bibitem[{Zeldes(2017)}]{zeldes2017gum}
Amir Zeldes. 2017.
\newblock \href {https://doi.org/10.1007/s10579-016-9343-x} {The gum corpus:
  creating multilayer resources in the classroom}.
\newblock \emph{Language Resources and Evaluation}, 51:581--612.

\bibitem[{Zhou et~al.(2019)Zhou, Khashabi, Ning, and
  Roth}]{zhou-etal-2019-going}
Ben Zhou, Daniel Khashabi, Qiang Ning, and Dan Roth. 2019.
\newblock \href {https://doi.org/10.18653/v1/D19-1332} {{``}going on a
  vacation{''} takes longer than {``}going for a walk{''}: A study of temporal
  commonsense understanding}.
\newblock In \emph{Proceedings of the 2019 Conference on Empirical Methods in
  Natural Language Processing and the 9th International Joint Conference on
  Natural Language Processing (EMNLP-IJCNLP)}, pages 3363--3369, Hong Kong,
  China. Association for Computational Linguistics.

\end{thebibliography}
\bibliographystyle{acl_natbib}
\clearpage
\appendix

\section{Numerical Stability of Relative Consistency}
    To avoid numerical instability, especially when comparing $\newconsistency$ scores for two models, (i.e. to determine whether a training approach improves a model's tendency to produce consistent responses, or determine which of two training approaches best improves a model's tendency towards consistent responses), we define:
        \begin{equation}
            \mu(c,a) = \sum_{\substack{c_i \in C_a \\ c_i \leq c}}{m(c_i,a)}
        \end{equation}
        (i.e., the cumulative combinatoric mass) and then rephrase the definition of $\newconsistency$ as:
        \begin{equation}
            \newconsistency(c,a)= \frac{\mu(c,a)}{M(a)}
        \end{equation}
        which relies on only one division, so is less prone to floating-point rounding errors.
        
        This also allows us to compute:
        \begin{equation}
           \frac{\mu(c_1,a_1)}{M(a_1)} - \frac{\mu(c_2,a_2)}{M(a_2)} 
        \end{equation}
        (i.e., the improvement in $\newconsistency(c_1, a_1)$ over $\newconsistency(c_2, a_2)$ scores) as:
        \begin{equation}
            \frac{\mu(c_1,a_1)M(a_2) - \mu(c_2,a_2)M(a_1)}{M(a_1)M(a_2)}
        \end{equation}
        which allows for comparisons between models that are very close in their $\newconsistency$ scores, (i.e., in the long tail of consistency).
    
\section{Formulas for Bundle Sizes $b > 2$}
    \label{appendix:b}
    Let us consider a contrastive test set containing $n$ bundles of $b$ instances each. There are $nb+1$ possible accuracies $a$, but still $n+1$ possible consistencies $c$.

        $C_a$  can then be defined in terms of $n$, $b$, and $a$ as follows:
            \begin{equation}
                C_a = \{c \in C : c_{min}^{(a)} \leq c \leq c_{max}^{(a)} \}
            \end{equation}
            where $c_{min}^{(a)}$ and $c_{max}^{(a)}$ are defined as:
            \begin{equation}
                c_{min}^{(a)} = 
                    \begin{cases}
                        0 & \text{if } a \leq n(b-1) \\
                        a-n(b-1) & \text{if } a > n(b-1
                    \end{cases}
            \end{equation}
            \begin{equation}
               c_{max}^{(a)} = \left\lfloor \frac{a}{b} \right\rfloor
            \end{equation}

        Intuitively, if  $a \leq n(b-1)$ then it is possible that all bundles have at least one of their constituent instances incorrectly answered, in which case, $c_{min}^{(a)} = 0$. However, if $a > n(b-1)$, then at least $a - n(b-1) > 0$ of bundles must be fully correctly answered. Indeed, for a bundle to be inconsistent at least one item must be incorrectly answered, so for a given $a$, the number of incorrect items is $nb-a$. Thus, at most $nb-a$ bundles can be inconsistent, and $c_{min}^{(a)} = n-(nb-a)=n-nb+a=a-n(b-1)$. 
        
        The definition of $c_{max}^{(a)}$ follows from the observation that a maximally consistent model will consistently respond to the maximum number of bundles for which it is possible that all $b$ instances are correctly answered, and that equals $\left\lfloor \frac{a}{b} \right\rfloor$.

        Now,  $M(a)$ (the number of ways a model can achieve accuracy $a$) is given by:
        \begin{equation}
            M(a) = {nb \choose a}
        \end{equation}
        and $m(c, a)$ (the number of ways a model can achieve accuracy $a$ and consistency $c$) is given by:
        \begin{equation}
            m(c, a) = {n \choose c} \cdot G(n-c, b, a - cb)
        \end{equation}
        where the first factor in the product still intuitively corresponds to the number of ways that $c$ consistent bundles can be selected out of $n$, but the second refers to the number of ways the remaining correct instances could be distributed within responses to the test set such that no additional consistent bundles can be formed. 
      
        This second factor $G(m,b,k)$ is defined as:
        \begin{equation}
            G(m,b,k) = \sum_{r=0}^{R}{\left(-1\right)^{r}{m \choose r}{(m-r)b \choose k-rb}}
        \end{equation}
        where $R = \min(m, \left\lfloor \frac{k}{b} \right\rfloor)$. This can be understood as the number of ways to select $k$ elements of an $m \times b$ matrix such that no row contains a complete $b$ elements selected. The first term (which simplifies to ${mb \choose k}$) is the total number of ways these $k$ elements could be selected, ignoring the restriction on complete rows, and the remaining terms apply the principle of inclusion-exclusion to alternately subtract and add the number of ways that at least $r$ rows could be filled (by multiplying the number of ways that $r$ out of $m$ rows could be selected by the number of ways the remaining $m-r$ rows and $b$ columns could be filled by the remaining $k-rb$ items to select), up to the maximal number of rows $R$ that could be filled, whether that is determined by the total number of rows available $m$ or the number of rows the items $k$ could fill.

        In general, we do not recommend using this measurement for bundle sizes above 2 except for evaluating consistency on three-valued features, as many consistency questions can be formulated as bundles with one instance exhibiting a feature and one instance lacking that feature.
        
\section{Distributions of Alternative Approaches}
    \label{appendix:other_distributions}
\begin{figure*}[!ht]
    \centering
    \includegraphics[width=.495\textwidth]{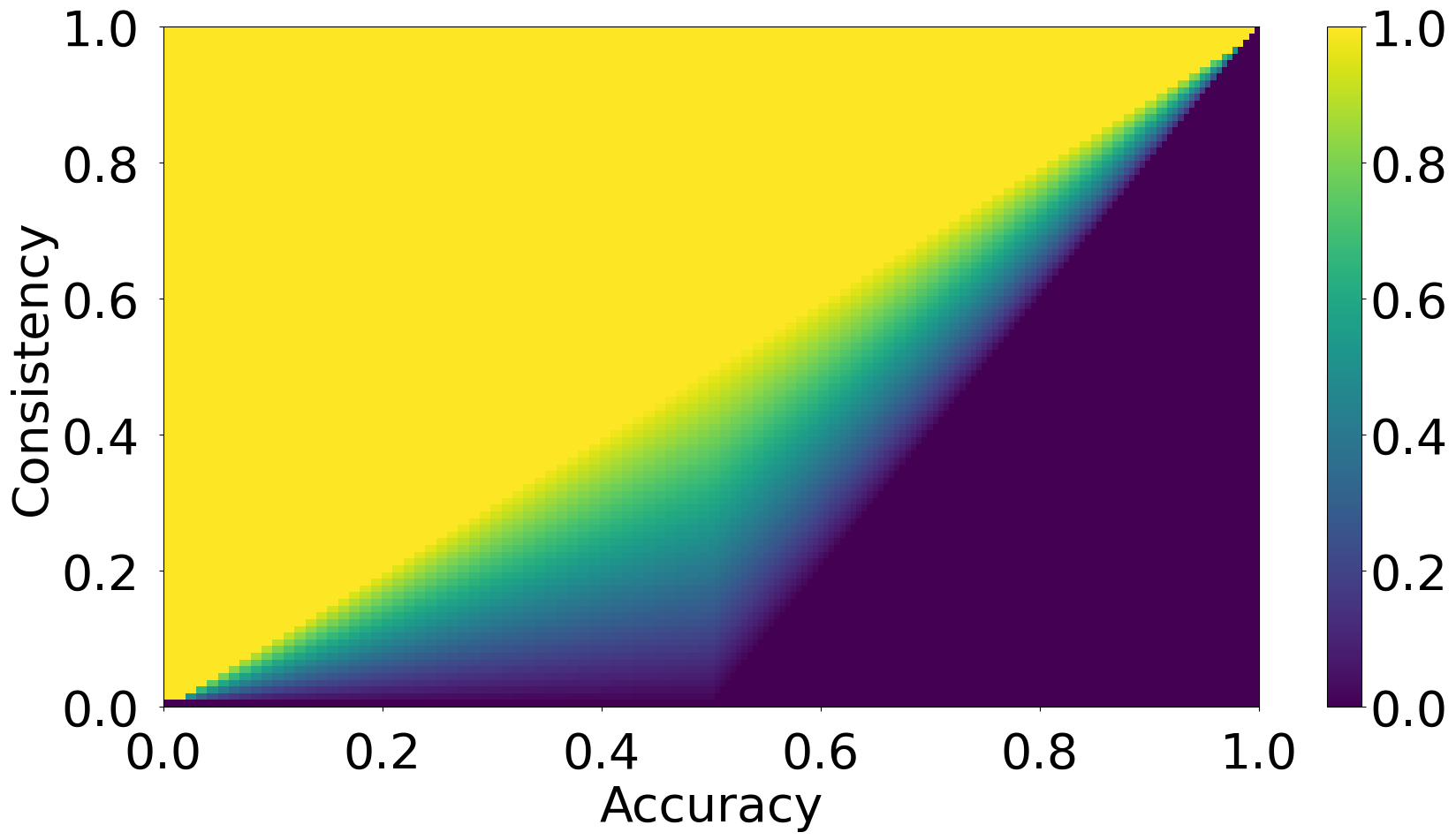}
    \includegraphics[width=.495\textwidth]{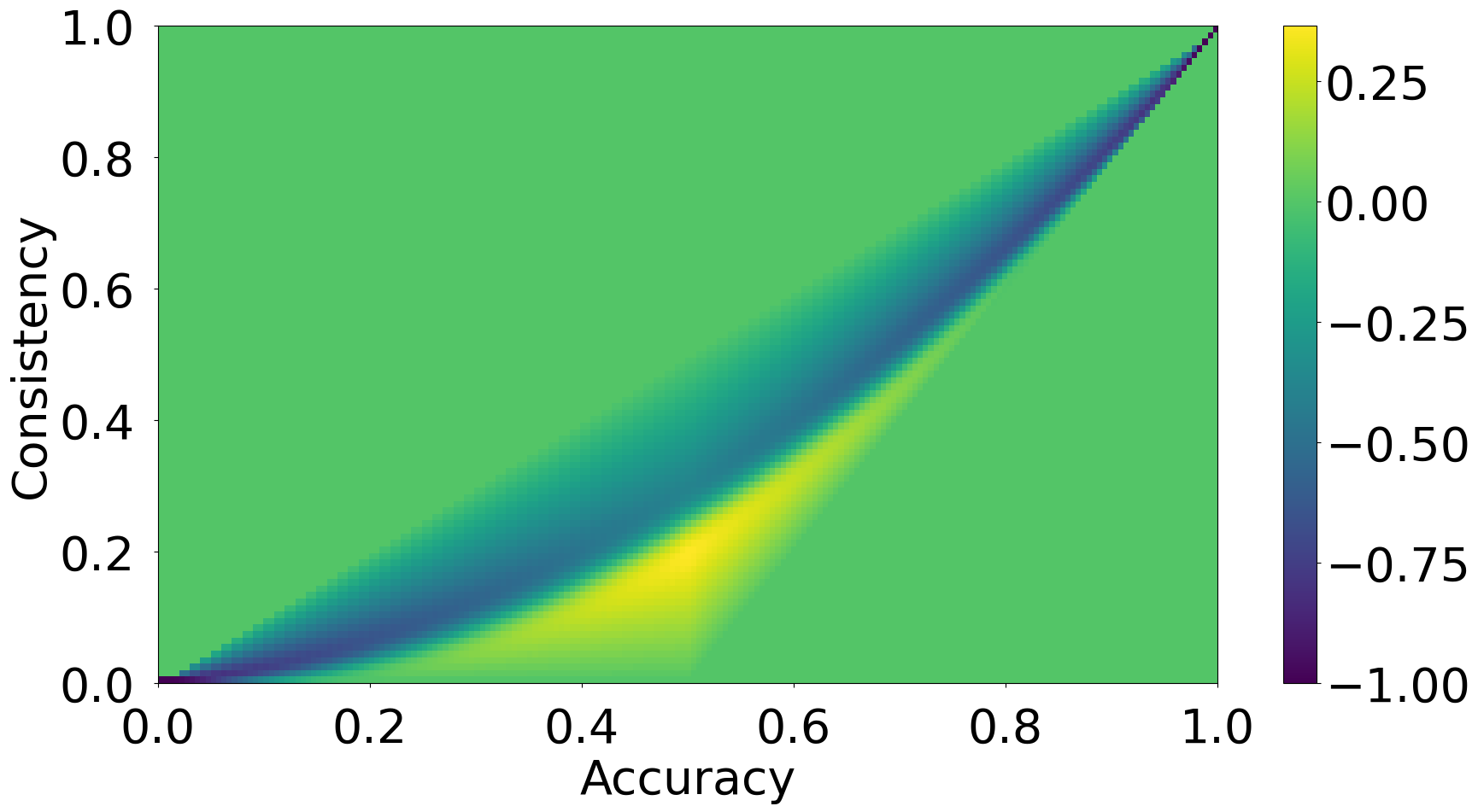}
    \caption{
        In this figure, the interval $\left[ c_{min}^{(a)}, c_{max}^{(a)} \right]$ is simply scaled to cover $[0,1]$ and the score is scaled accordingly. On the left is the score given a model's accuracy and consistency, on the right is shown the change in score when moving from $\newconsistency$ to this formulation.
    }
    \label{fig:r_distribution}
\end{figure*}

\begin{figure*}[!ht]
    \centering
    \includegraphics[width=.495\textwidth]{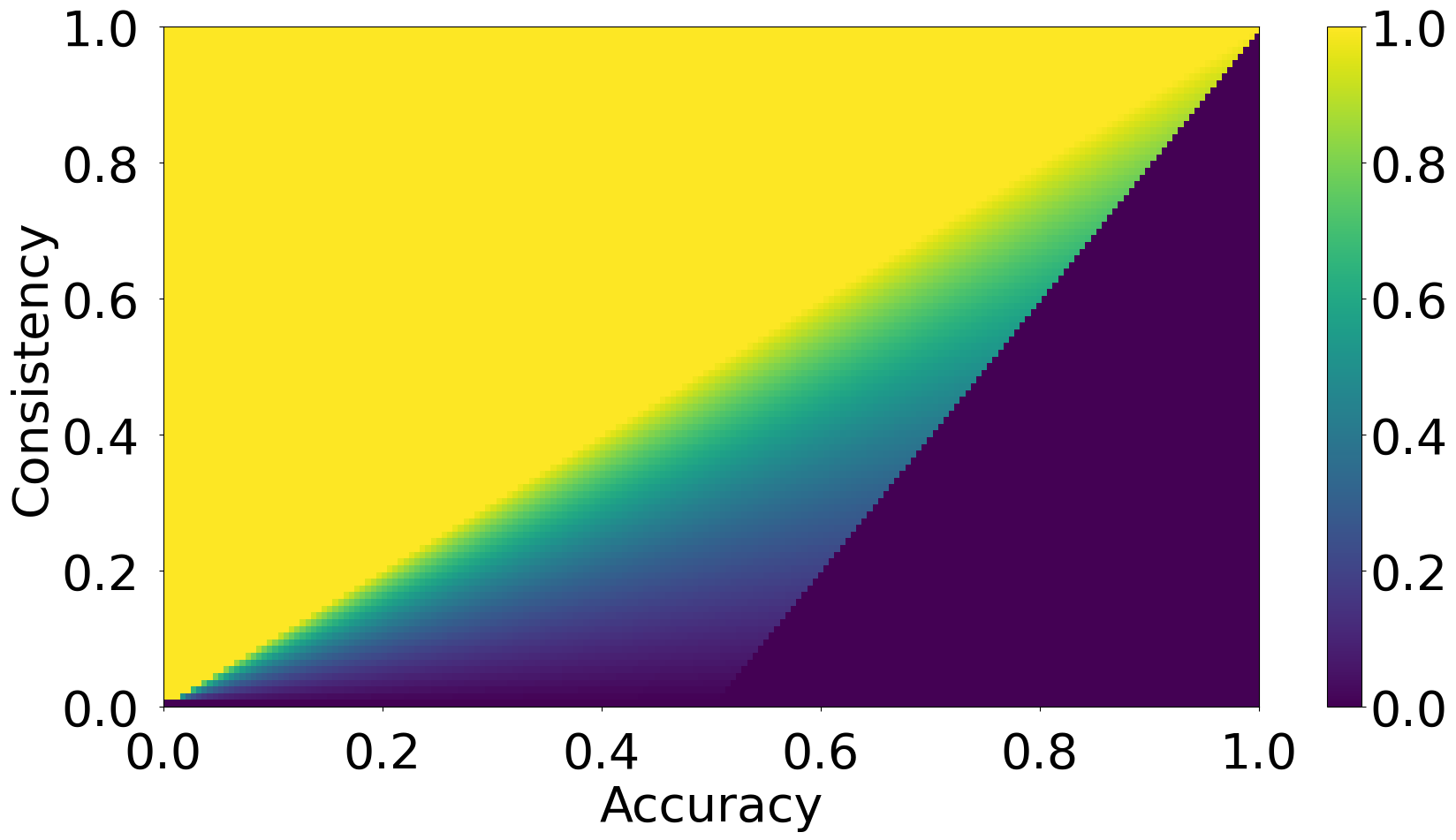}
    \includegraphics[width=.495\textwidth]{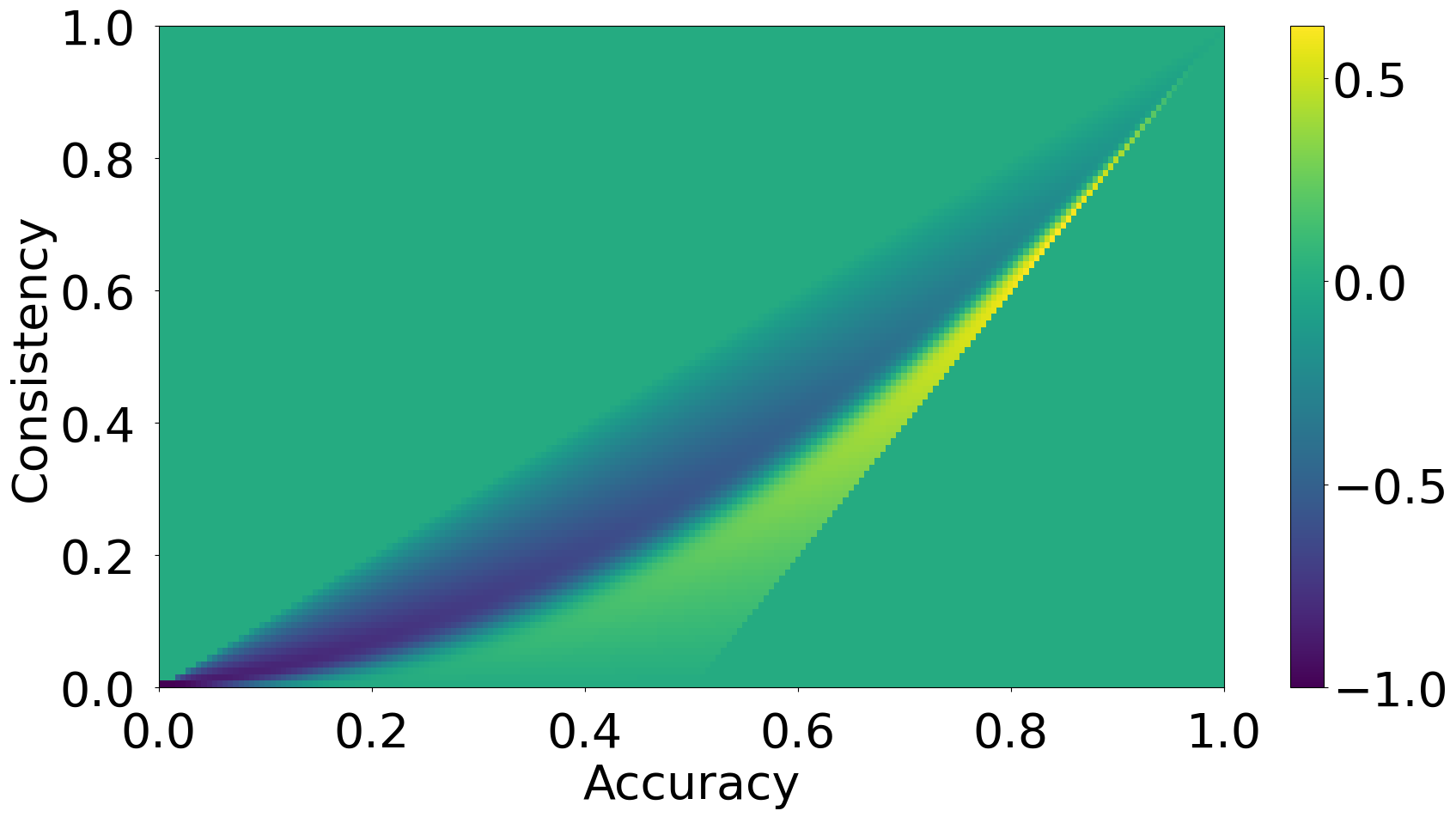}
    \caption{
        In this figure, of the bundles which are at least partially correct, the proportion of fully consistent bundles is reported. On the left is the score given a model's accuracy and consistency, on the right is shown the change in score when moving from $\newconsistency$ to this formulation.
    }
    \label{fig:r2_distribution}
\end{figure*}

    Figures \ref{fig:r_distribution} and \ref{fig:r2_distribution} plot the distributions of consistency scores (for a 100-bundle dataset) obtained via simpler non-probabilistic alternatives and compare them to the distributions obtained via $\newconsistency$. 
    Both of these characterizations lower the scores for consistencies that are above chance and raise the scores for consistencies that are below chance. 

\begin{figure}[t]
    \centering
    \includegraphics[width=.495\textwidth]{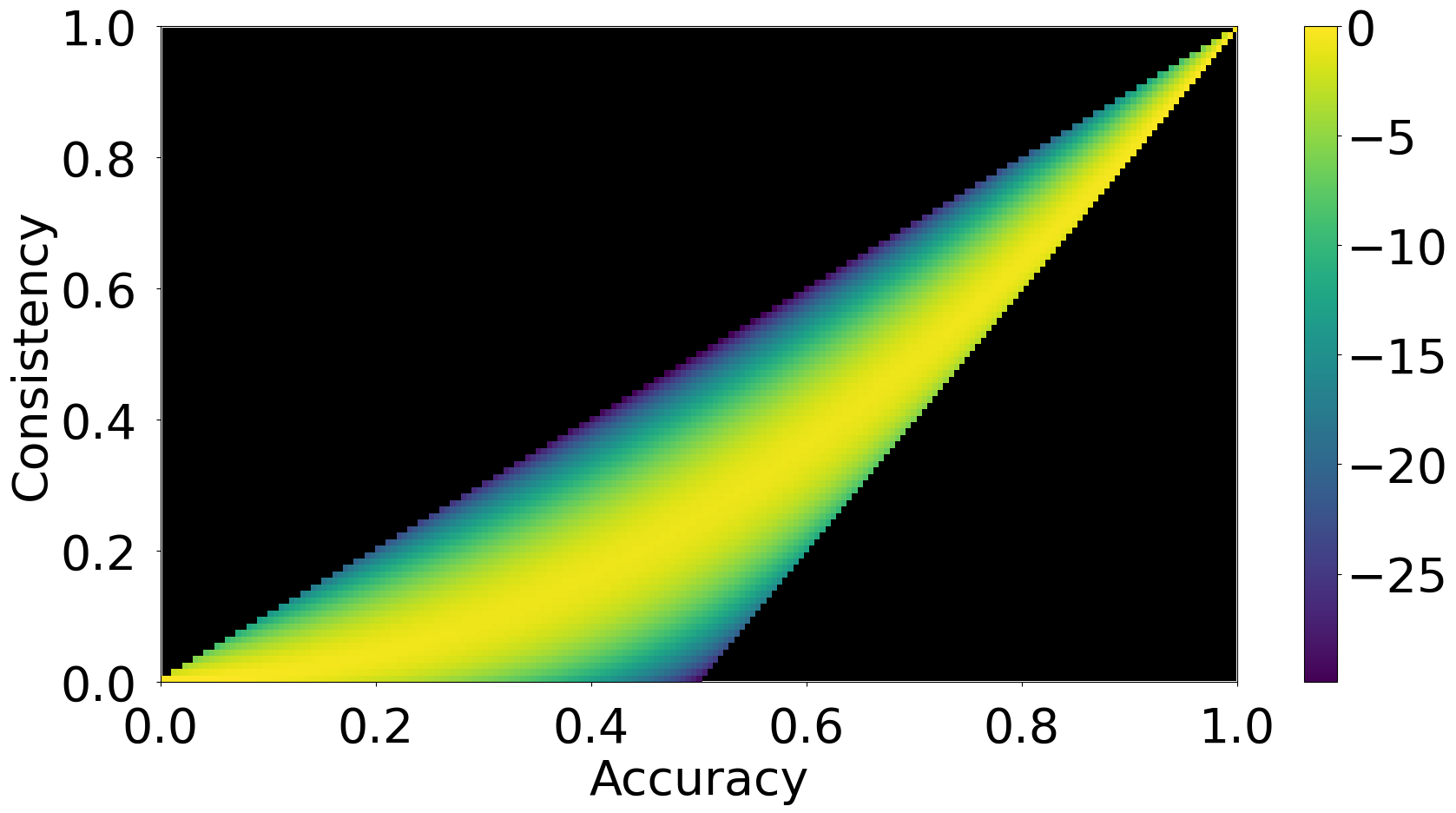}
    \caption{
        The $\log_{10}$ of the distributions of consistency scores in Figure \ref{fig:c_distribution_a}. 
    }
    \label{fig:distribution_log10}
\end{figure}
\begin{table*}[t]
\resizebox{\textwidth}{!}{
\begin{tabular}{llrrrrrrrrrrrr}
\toprule
\textbf{Size} & \textbf{Loss} & \textbf{B-A} & \textbf{P-A} & \textbf{S-A} & \textbf{A-A} & \textbf{B-C} & \textbf{P-C} & \textbf{S-C} & \textbf{A-C} & \textbf{B-RC} & \textbf{P-RC} & \textbf{S-RC} & \textbf{A-RC} \\
\midrule

\multirow{5}{*}{\rotatebox{90}{Large}} 
    & MLE &
        67.22 & 66.33 & 66.84 & 67.09 &  27.04 & 58.16 &  42.86 &                       42.35  & 99.92 & 100.00 &  17.10 &                       10.06  \\
    & CE &
        67.35 & 67.35 & 64.80 & 66.84 &                       28.57  & 61.22 &                        40.31 & 43.37 &                       99.99  & 100.00 &                       20.10  & 26.64 \\
    & $\lambda_1$MLE + $\lambda_2$CE\\
    & \quad \rotatebox[origin=c]{180}{$\Lsh$} $\lambda_1, \lambda_2=1.0,1.0$ & 
        67.73 & 68.88 & 66.33 & 68.11 &                       28.57  & 63.78 &  42.86 & 44.39 &                       99.98  & 100.00 &  30.43 & 19.37 \\
    & \quad \rotatebox[origin=c]{180}{$\Lsh$} $\lambda_1, \lambda_2=0.33,1.0$ &
        68.24 & 68.37 & 66.58 & 68.37 &                       30.10  & 63.27 &                       43.37  &                       44.90  &                       100.00 & 100.00 &                       42.44  &                       16.01  \\

\midrule

\multirow{5}{*}{\rotatebox{90}{3B}} 
    & MLE &
        75.64 & 76.28 & 74.23 & 76.79 &                       44.39  & 71.43 & 56.12 &                       60.71  &                       100.00 & 100.00 & 80.76 &                       88.68  \\
    & CE  &
        75.38 & 75.51 & 74.23 & 77.55 & 43.88 & 70.41 & 56.12 & 61.73 & 100.00 & 100.00 & 80.76 & 92.03 \\
    & $\lambda_1$MLE + $\lambda_2$CE\\
    & \quad \rotatebox[origin=c]{180}{$\Lsh$} $\lambda_1, \lambda_2=1.0,1.0$ & 
        75.51 & 75.77 & 74.23 & 77.04 & 44.90 & 70.92 & 56.12 & 60.71 & 100.00 & 100.00 & 80.76 & 88.68 \\
    & \quad \rotatebox[origin=c]{180}{$\Lsh$} $\lambda_1, \lambda_2=0.33,1.0$ &
        76.53 & 77.55 & 76.02 & 78.57 &                       45.92  & 73.47 & 58.67 &                       63.78  &                       100.00 & 100.00 & 79.32 &                       98.60  \\                          \bottomrule

\end{tabular}
}
\caption{The full results of UnifiedQA-v2 \cite{khashabi2022unifiedqav2} on the CondaQA contrastive dataset, with the expectation that including the Contrastive Estimation (CE) objective would improve consistency, as in \citet{dua-etal-2021-learning}.}
\label{tab:ravichander_et_al-appendix}
\end{table*}
\end{document}